\documentclass[sigconf]{aamas}  
\usepackage{balance}  

\usepackage{booktabs}

\usepackage{booktabs}
\usepackage[T1]{fontenc}
\usepackage{lmodern}
\usepackage{multirow}
\usepackage{amsfonts}
\usepackage{amsmath,amssymb,amsfonts}

\usepackage{algorithmic}
\usepackage{textcomp}
\usepackage{xcolor} 
\usepackage[linesnumbered, lined, boxed, ruled, vlined]{algorithm2e}
\usepackage{verbatim}
\usepackage{enumitem}

\setcopyright{ifaamas}  
\acmDOI{}  
\acmISBN{}  
\acmConference[AAMAS'19]{Proc.\@ of the 18th International Conference on Autonomous Agents and Multiagent Systems (AAMAS 2019)}{May 13--17, 2019}{Montreal, Canada}{N.~Agmon, M.~E.~Taylor, E.~Elkind, M.~Veloso (eds.)}  
\acmYear{2019}  
\copyrightyear{2019}  
\acmPrice{}  

\begin{document}

\title{Can Sophisticated Dispatching Strategy Acquired by Reinforcement Learning? }  
\subtitle{A Case Study in Dynamic Courier Dispatching System}


\author{Yujie Chen}
\affiliation{%
  \institution{Zhejiang Cainiao Supply Chain Management Co., Ltd}
}
\email{aisling.cyj@cainiao.com}
\author{Yu Qian}
\affiliation{%
  \institution{Zhejiang Cainiao Supply Chain Management Co., Ltd}
}
\email{qianyu.qy@alibaba-inc.com}
\author{Yichen Yao}
\affiliation{%
  \institution{Zhejiang Cainiao Supply Chain Management Co., Ltd}
}
\email{eason.yyc@alibaba-inc.com}
\author{Zili Wu}
\affiliation{%
  \institution{Zhejiang Cainiao Supply Chain Management Co., Ltd}
 }
\email{zili.ziliwu@cainiao.com}
\author{Rongqi Li}
\affiliation{%
  \institution{Zhejiang Cainiao Supply Chain Management Co., Ltd}
}
\email{rongqi.lrq@cainiao.com}
\author{Yinzhi Zhou}
\affiliation{%
  \institution{Zhejiang Cainiao Supply Chain Management Co., Ltd}
}
\email{yinzhi.zyz@cainiao.com}
\author{Haoyuan Hu}
\affiliation{%
  \institution{Zhejiang Cainiao Supply Chain Management Co., Ltd}
}
\email{haoyuan.huhy@cainiao.com}
\author{Yinghui Xu}
\affiliation{%
  \institution{Zhejiang Cainiao Supply Chain Management Co., Ltd}
}
\email{renji.xyh@taobao.com}

\begin{abstract}  
	In this paper, we study a courier dispatching problem (CDP) raised from an online pickup-service platform of Alibaba. The CDP aims to assign a set of couriers to serve pickup requests with stochastic spatial and temporal arrival rate among urban regions. The objective is to maximize the revenue of served requests given a limited number of couriers over a period of time. Many online algorithms such as dynamic matching and vehicle routing strategy from existing literature could be applied to tackle this problem. However, these methods rely on appropriately predefined optimization objectives at each decision point, which is hard in dynamic situations. This paper formulates the CDP as a Markov decision process (MDP) and proposes a data-driven approach to derive the optimal dispatching rule-set under different scenarios. Our method stacks multi-layer images of the spatial-and-temporal map and apply multi-agent reinforcement learning (MARL) techniques to evolve dispatching models. This method solves the learning inefficiency caused by traditional centralized MDP modeling. Through comprehensive experiments on both artificial dataset and real-world dataset, we show: 1) By utilizing historical data and considering long-term revenue gains, MARL achieves better performance than myopic online algorithms; 2) MARL is able to construct the mapping between complex scenarios to sophisticated decisions such as the dispatching rule. 3) MARL has the scalability to adopt in large-scale real-world scenarios.
\end{abstract}

%


\maketitle


\section{Introduction}
With the rapid development of e-commerce, millions of small online retailers have emerged, resulting in a large growth in demand for pickup service. Only if the couriers pick up the goods within the prescribed time window, the subsequent transportation and delivering service can be carried out timely. 

In this study, we consider a courier dispatching problem (CDP) with dynamic customers. In particular, given a number of couriers and unserved requests, decision should be made to assign couriers to pickup requests that maximize the total revenue. Each request has a hard service time window and the information is known when the system receives the customer's request. 

\begin{figure}[ht]
	\center
	\includegraphics[width=\linewidth]{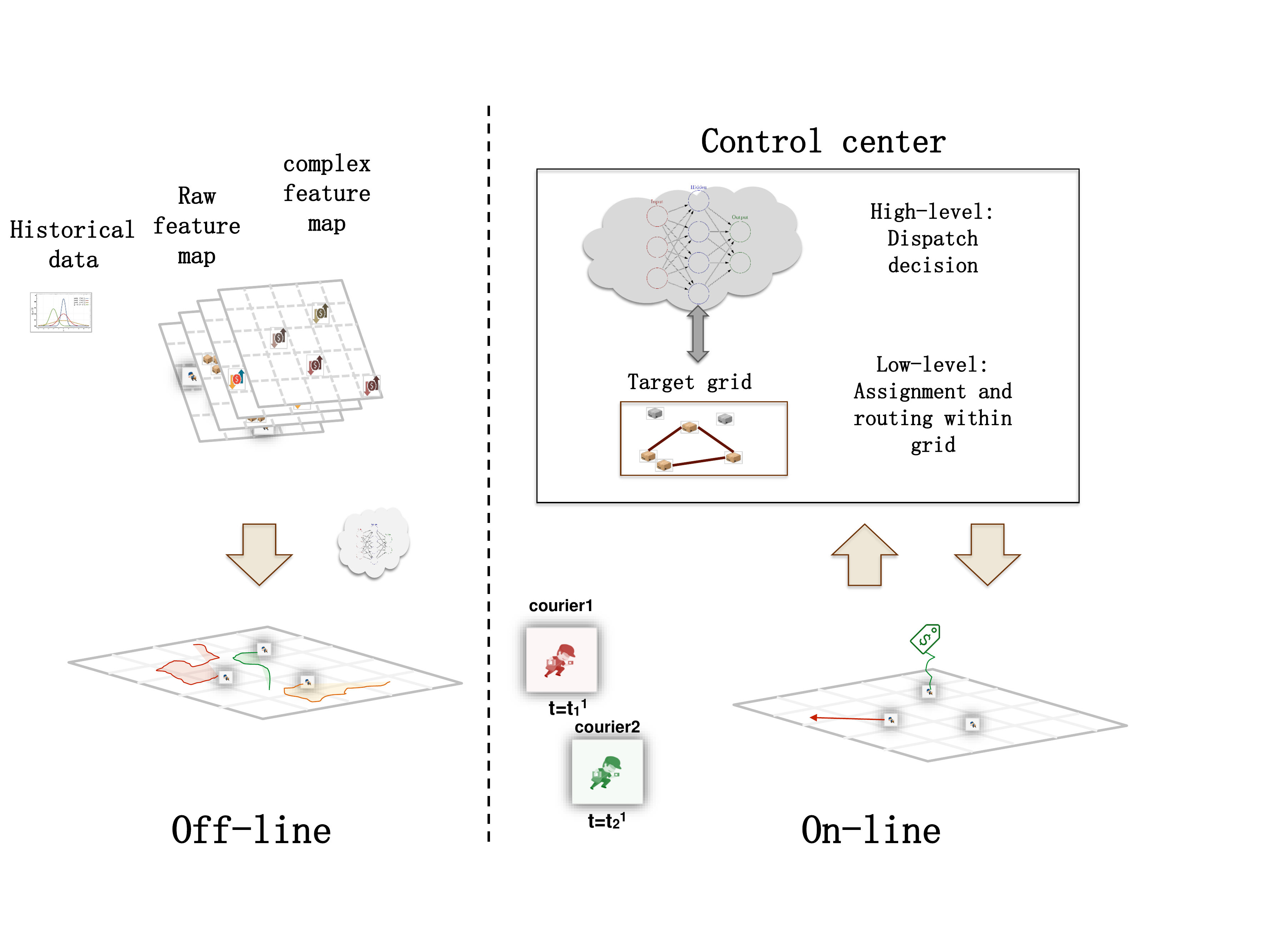}
	\center
	\caption{Overview of the proposed decision mechanism}
	\label{fig:Overview}
\end{figure}

A straightforward solution for the CDP is to formulate and solve an optimization problem with the development of the problem. However, these methods rely on appropriately predefined optimization objectives at each decision point, which is hard in dynamic situations. To capture the stochastic spatial-and-temporal pattern of pickup requests and utilize the information in the decision process, we discretize the urban region into cartesian grids. Then, a hierarchical solution approach is designed to handle these pickup requests, namely ``dispatching-level'' and ``routing-level''. At the ``dispatching-level'', we adopt a learning model to uncover the spatial-and-temporal pattern of customer demands. The dispatching model aims to find an optimal policy that designates each courier to a target grid and at the same time assigns a service time for each courier in his/her target grid. At the ``routing-level'', the problem within the target grid is formulated as an orienteering problem with time windows \cite{Kantor1992}. Briefly, given a set of geographically distributed customers with different prices and time windows, a vehicle aims to maximize its income within limited duration constraint. To solve this problem, we adopt a deterministic insertion algorithm proposed by \cite{Kantor1992}. Figure \ref{fig:Overview} depicts our overall decision mechanism.

This paper focuses on the study of learning models at the ``dispatching-level''. In recent years, multi-agent reinforcement learning (\emph{MARL}) techniques have been successfully applied to a few traditional operational research problems, such as supply chain management \cite{Stockheim2002}, inventory optimization \cite{Oroojlooyjadid2017} and fleet management problem \cite{Lin2018}. Although the information quality is improved by many service infrastructures, it still remains significant challenges when adopting \emph{MARL} approach on CDP. Similar to the fleet management problem stated in \cite{Lin2018}, the CDP also faces the difficulties of 1) problem setting designs including state-action space and appropriate reward functions, 2) large numbers of agents issue and 3) the cooperation problem between agents. In order to provide more realistic decisions, our \emph{MARL} is evaluated in a more complex environment that:

\begin{enumerate}[leftmargin=*]
    \item Each pickup request has more complex behavior information such as arrival time, earliest start time, latest start time and service time. Consequently, we can not use a simple linear function to derive the corresponding reward for each action.
    \item Couriers are assigned with tasks whenever it becomes free, and get the actual reward when it fulfills the requests in grids. This setting leads to a delayed reward. Due to the long processing time of each action, mutual influence between agents is magnified, which increases the difficulty in the design of reasonable feedback mechanism.
\end{enumerate}

Our major contributions are as follows:

\begin{itemize}[leftmargin=*]
	\item It is the first time to solve the CDP using multi-agent reinforcement learning method based on Markov decision process.
	\item We propose an effective dispatching algorithm with decentralized control and reward shaping to introduce cooperation between agents.
	\item Our proposed method outperforms a set of well-studied online algorithms for CDP through comprehensive experiments on both of the artificial dataset and the real-world dataset. Moreover, the generalization of our method is confirmed on unseen scenarios beyond the training dataset, mainly on scope of customer time window, dynamic ratio and zone distribution variation.    
\end{itemize}

\section{Related Work}
In this section, we discuss traditional models and solutions to solve the CDP. Additionally, we also present multi-agent reinforcement learning approaches that recently appeared in related problems. 
 
\subsection{Models for CDP}
The courier dispatching problem (CDP) is commonly considered as a dynamic variant of the vehicle routing problem (DVRPs) with time windows \cite{Sungur2010}, or an online matching problem \cite{Spivey2004}.  According to the type of the uncertain information, studies have been carried out in uncertain customer demands \cite{Angelelli2005,Goodson2013}, stochastic travel time \cite{Laporte1992,Yan2013}, stochastic service time \cite{Sungur2010} and customer locations \cite{Ulmer2016}. However, the common thing is that all these models tackles the dynamic and stochastic characteristic of the CDP.

\subsection{Heuristics for CDP}
To solve these dynamic problems, we categorize existing approaches based on whether any prediction of future events is utilized in the decision process, namely the myopic strategy and the looking-forward strategy. 

In the first group, the solution process is based on known data and does not consider any uncertain information. Ritzinger et al. pointed out that it is hard to compromise between reactiveness and decision quality in dynamic scenarios \cite{Ritzinger2016}. The challenge of these myopic solutions is to set up appropriate objective functions. Heuristics from static scenarios can be modified for dynamic situations, such as tabu search \cite{Gendreau2006} and large neighborhood search \cite{Hong2012}.

Researches from many real-world applications show the underline data pattern of customer demands \cite{Malandraki2001}. The looking-forward strategy predicts future events either implicitly or explicitly. Sungur et al. use stochastic programming with recourse to model the uncertainty in demands \cite{Sungur2010}. During the offline phase, a robust optimization is proposed to derive a master plan. During the operation phase, their recoursing rule simply omits non-occurring customers and reschedules new customers. A different approach is to embed sampling techniques into static solvers \cite{Sarasola2016}. 
The most similar work to our research is to model the CDP using the language of Markov decision processes (MDP). For any practical sized DVRPs, obtaining optimal policies is computationally intractable \cite{Dror1989}. Thus, heuristic methods with rollout policy are commonly applied \cite{Novoa2009,Goodson2013,Ulmer2016}. More practically, a dynamic lookup table technique that simplifies and discretizes the states during the optimization is proposed recently \cite{Ulmer2017,Ulmer2018}.

\subsection{Multi-agent RL (\emph{MARL}) for planning}
There are many obstacles to adopt multi-agent reinforcement learning(\emph{MARL}) algorithms in real-world applications, which often involves complicated interactions between multiple agents. Furthermore, feedback from the environment is also non-linear. Further difficulties include non-stationary environment and prohibitively large and intractable state-action space \cite{Lin2018}. With the development of \emph{MARL} and deep learning research, multi-agent systems have been recently studied in a variety of domains including robotic teams and resource management \cite{Busoniu208,Bakker2010}. In a \emph{MARL} setting, it is usually challenging to specify a good objective function, since the returns of agents are correlated \cite{Busoniu2010}. By applying experience sharing technique, homogeneous agents can learn faster and reach better overall performance \cite{Stankovic2016}. Recently, Oroojlooyjadid et al. adopted deep Q-Learning to optimize the replenishment decisions at a given stage in a supply chain management application \cite{Oroojlooyjadid2017}. More similar to our work, Lin et al. proposed a contextual multi-agent actor-critic(CMAAC) method to find the optimal dispatching rules that balance the supply and demand in a geographical area \cite{Lin2018}. However, the size of CMAAC's action space is limited to 6 neighboring girds to move, which is a less optimal decision for real-world online dispatching system.

\section{Problem Statement and Simulator Design}
\label{sec:problem}     
\paragraph{Problem}
Our study is carried out on a $20\times20$ grid world, where pickup requests arrive over time. Each pickup request specifies the earliest start time, the latest start time and the service time. We aim to design a dispatching algorithm that maximizes the total revenue of all couriers within limited working hours. In particular, the dispatching algorithm generates two parts of decisions: 1) assigning the courier to a target grid; 2) and assigning a service time for the courier in the target grid at the same decision time. 

\paragraph{Simulator}
To evaluate algorithm performance, we introduce a simulator that generates the pickup requests and executes the decisions of the tested dispatching algorithm. All the later experiments are carried out on this simulator. 

\begin{figure}[ht]
	\center
	\includegraphics[width=\linewidth]{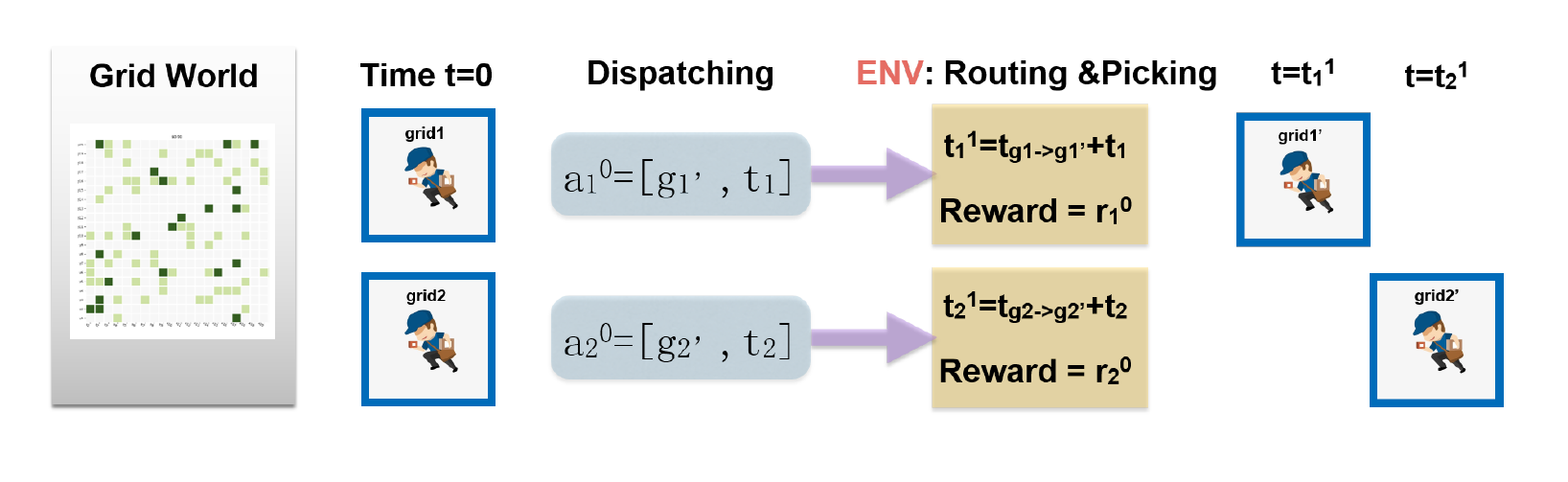}
	\center
	\caption{Environment Settings}
	\label{fig:Environment}
\end{figure}

Our simulator applies a discrete event driven model. At the start time, all couriers send a request to the dispatching algorithm and receive an instruction. As long as any courier completes his/her current task, the dispatching algorithm assigns a new task to the courier. Whenever a courier arrives his/her target gird, a deterministic insertion algorithm (DIA)\cite{Kantor1992} is called to give the detailed pick-up instructions. The input parameters of DIA include the target gird arrival time, allowed service time and real-time customers at the arrival time of the target grid. In particular, as shown in Figure \ref{fig:Environment}, the ``routing-level'' decision is treated as a part of the simulator. When a courier completes all assigned requests, the corresponding revenue and the actual task execution time is recorded for this action and the following activities are conducted sequentially:

\begin{itemize}[leftmargin=*]
	\item \emph{Order generation:} new pickup requests are dynamically added to the environment as soon as the new request appears, and are removed from the environment when current time goes beyond the customer service time window. Hiring more couriers can reduce the portion of not served requests, and this relates to the tradeoff between the labor cost and platform revenue. In the environment setting, the number of couriers can fulfill above 80\% of the requests in a whole day, which is at a reasonable level for performance comparison of different algorithms.
	
	\item \emph{Courier status updates: } each courier has three status defined as follows: 
	 \subitem 1) free: ready for the next instruction
	 \subitem 2) walking: on the way to target grid
	 \subitem 3) picking: executing the pickup services within the target grid.
\end{itemize}

\section{Learning for Dispatching}
In this section, we propose a Multi-Agent Reinforcement Learning (MARL) method to solve the CDP.
\subsection{MDP Definition}
\label{MDP Definition}
We build the Markov decision process (MDP) from a decentralized point of view that each courier is modeled as an agent. By using the decentralization language, the size of the state-action space is much controllable. The centralized model leads to an exponentially increased size of the state-action space when the number of courier increases. It is worth mentioning that even in a decentralized perspective, global information such as the generation of pickup requests, the status of other agents can be perceived. The other components of the MDP are defined as follows: 

\begin{itemize}[leftmargin=*]
	\item \textbf{State} $s_t^i \in \mathcal{S}$: The state of courier $i$ at time step $t$ considers informations in the $9\times9$ neighboring grids of the courier $i$. We use image-like tensor input as state, with each channel containing specific information about the environment. Two sets of states are discussed in the experiment, which differs in the information level the dispatch system could observe. At basic level, the dispatch system obtain the current snapshot of customers and couriers. The states include the number of agents, the number of pickup requests and the total price the requests in each grid, plus the distance of each neighbor grid to the current grid. At pre-solved level, dispatch system get extra information by calling the route planning solver. The presumed score of all possible actions at the current time is added as new channels. 
	
	\item \textbf{Action} $a_t^i \in \mathcal{A}$: At the ``dispatching-level'', the decision includes the target grid and the maximum patrol time within the grid. Therefore, our action space is the cartesian product of candidate grids and time periods. We restrict the candidate grids to the $5\times5$ neighborhood grids of the courier $i$. And the maximum patrol time is discretized into $\{0,10,20,30\}$ minutes. To avoid agents moving beyond the gird board, the target grid is clipped at the edge of the board. All the possible actions are represented as one-hot encoding. During the training stage, an action is chosen by roulette wheel selection according to probability distribution given by the model. During the testing stage, the action with the highest probability is selected.
	
	\item \textbf{Reward Function} $r_t^i \in \mathcal{R} =  \mathcal{S} \times  \mathcal{A} \rightarrow \mathbb{R}$: The reward design determines the optimization goal of the trained model. Intuitively, the reward for courier $i$ at time step $t$ is defined as total price of served requests according to last instruction. Moreover, the cooperation between couriers can be introduced through reward shaping, putting a weight $\alpha$ on how much each courier should care about its individual reward function versus the average of reward of all couriers. For example, 
	\begin{equation}\label{formula:shaping}
		r_{t, shaping}^i = r_t^i + \alpha \times R_{team}
	\end{equation}	

	The effectiveness of considering team work is empirically discussed in later sections.
	
	\item \textbf{Agent}: A decentralized policy is applied as the homogeneous property of couriers. All the agents establish a common policy through parameters sharing. During the execution phase, each courier is modeled as an agent that selects an action based on its own observations mentioned above.
	
	\item \textbf{Discount factor}: The discount factor controls the degree of how far the MDP looks into the future. In our application, the state transitions are affected by the joint behaviors of all agents. Meanwhile, the long processing time of each action introduces large uncertainty during the training process.  Therefore, a relatively small discount factor is preferable compared with the single-agent scenario. In our setting, the discount factor ${\gamma}=0.8$ is adopted. 
\end{itemize}

\subsection{MARL}
This section presents a Multi-Agent Reinforcement Learning based on actor-critic framework. There are two main ideas in the design of \emph{MARL}: 1) Decentralized value function and policy network are used with an expected update; 2) Reward shaping introduces explicit cooperation between agents and considers the whole gains of dispatching system.

\textbf{Value Network}. The decentralized state-value function is learned by minimizing the following loss function derived from Bellman equation: 
\begin{eqnarray}\label{formula:value loss}
	&L_{\theta_v} = \left( V_{\theta_v} \left(s_t^i\right) - V_{target}\left(s_{t+1}^i; \pi \right) \right)^2
\end{eqnarray}
\begin{eqnarray}\label{formula:target def}
	V_{target}\left(s_{t+1}^i; \pi \right) = r_t^i +V_{\theta_v'} \left(s_{t+1}^i\right)
\end{eqnarray}

where we use $\theta_v$ to denote the parameters of the value network, $\pi$ to denote the dispatching policy and ${\theta_v'}$ to denote the target value network. In order to stabilize the learning process, we fix the target network for a few episodes sampling's time. Moreover, efficient corporation among agents can be established on this decentralized value network.

\textbf{Policy Network}. Policy gradient methods work by computing an estimator of the policy gradient and plugging it into a stochastic gradient ascent algorithm. The most commonly used gradient estimator has the form:
\begin{eqnarray}\label{formula:gradient est}
	g = \mathbb{E}_t\left[\nabla_{\theta} log_{\pi_{\theta}}\left(a_t | s_t\right)A_t\right] 
\end{eqnarray}
where $\pi_{\theta}$ is a stochastic policy and $A_t$ is an estimator of the advantage function as time step $t$, which is $A$($s_t, a_t$) for short. In this paper, we use the same objective proposed in the state-of-art actor-critic algorithm from Proximal Policy Optimization(PPO):
\begin{eqnarray}\label{formula:gradient ppo}
	L_{\theta} = \mathbb{E}_t \left[\min \left (r_t \left(\theta \right), clip \left(r_t \left(\theta \right),1-\epsilon,1+\epsilon \right) \right){A_t} \right]  
\end{eqnarray}
\begin{eqnarray}\label{formula:advantage}
	A_t = -V_{\theta_v} \left(s_t\right) + r_t + \dots + \gamma^{T-t+1} r_{T-1} + \gamma^{T-t}V\left(s_t\right)
\end{eqnarray}
where $\theta$ denotes the parameters of policy network and the expectation $\mathbb{E}_t$ indicates the empirical average over samples. The probability ratio $r_t$($\theta$) is computed as follows:
\begin{eqnarray}\label{formula:prob ratio}
	r_t \left(\theta \right) = \frac{ \pi _{ \theta } \left(a_t| s_t\right)} { \pi _{ \theta _{old}}\left( a_t| s_t\right)} 
\end{eqnarray}
Based on the MDP definition, we collect a set of tuples ($s, a, s', r$) based on our simulator environment mentioned in Section 3. Note that since we do not distinguish individual couriers, the collections of all couriers form a joint-dataset for learning value and actor network. The detailed description of \emph{MARL} is summarized in Algorithm \ref{algo:1}.

\begin{algorithm}
	\renewcommand{\algorithmicrequire}{\textbf{Input:}}
	\renewcommand\algorithmicensure {\textbf{Output:} }
	\caption{Multi-agent Training Framework}
	\begin{algorithmic}[1]
		\label{algo:1}       
		\STATE Initialize replay memory ${M}$
		\STATE Initialize actor net and critic net
		\FOR {{m = 1 to $N_{episode}$} }
		\STATE Random choose instance from train set
		\STATE Reset environment and get initial state
		\STATE \textbf{Stage 1: Sampling}
		\WHILE {\emph{t < T}}
		\STATE Sample actions \textbf{a}$_t$ according to policy network, given \textbf{s}$_t$
		\STATE Execute \textbf{a}$_t$ in the simulator and observe reward \textbf{r}$_t$, next state \textbf{s}$_{t+1}$
		\STATE Compute value network target as Eq(\ref{formula:target def}) and advantage as Eq(\ref{formula:advantage})
		\STATE Store the transitions ($s_t^i, a_t^i, r_t^i, s_{t+1}^i$) for all couriers $i \in $[$1,\dots,N$] into memory ${M}$       
		\ENDWHILE
		\STATE \textbf{Stage 2: Learning}
		\FOR {{$n_1$ = 1 to $N_1$} }
		\STATE Sample a batch of experience: $s_t^i$, $V_{target}$($s_{t+1}^i; \pi $)
		\STATE Update value network by minimizing the value loss Eq(\ref{formula:value loss}) over the batch
		\ENDFOR 
		\FOR {{$n_2$ = 1 to $N_2$} }
		\STATE Sample a batch of experience: $s_t^i$, $a_t^i$, $A$($s_t, a_t$)
		\STATE Update policy network as $\theta \leftarrow \theta + \nabla_{\theta} L_{\theta}$ according to Eq(\ref{formula:gradient ppo})
		\ENDFOR 
		\ENDFOR  
		
	\end{algorithmic}
\end{algorithm}

\section{Data Description}
In this section, we describe the instances of the simulation environment. We create artificial datasets to empirically test the performance of different algorithms, with the scenario settings varies \cite{Psaraftis2016}. In addition, we also apply the model to the real-world pickup-service data to verify the feasibility. 

\subsection{Time Horizon}
Our pick up services are provided within a fixed time horizon. In our settings, we set the time horizon to 480 minutes, which represents the working period from 8:00 am to 4:00 pm. Within this horizon, discrete pickup requests are generated following a given distribution. Couriers are not required to return to the depot at 4:00 pm, but they would not accept any new requests. The revenue of the day is the total price of all served requests.

\subsection{Temporal \& Spatial Distribution}
The arrival rate of new requests in a city varies over time and space. In our data generator, we divide the service area into $G$ grids and partition the time horizon into $M$ intervals. We use $\lambda_{gm}$ to denote the request arrival rate in grid $g$ at time interval $m$. In each time interval $m$, every grid $g$ generates a series of new arrival requests with the arrival rate of $\lambda_{gm}$ in a Poisson process. Given a new generated request $r_i$ generated at time $t_i$, the corresponding time windows is set as follows: 1) the earliest start time $\underline{w_i}$ is at $t_i+\Delta T_1$ , where $t_i$ is the arrival time of request $r_i$ ; 2) the latest that time $\overline{w_i}$ is set to $\underline{w_i}+\Delta T_2$. The parameters $\Delta T_1$ and $\Delta T_2$ are constant value. In addition, customer service time is uniformly chosen from the set \{2, 3, 4\} minutes, and a customer price is uniformly chosen from the set \{1, 2, 3, 4, 5\} dollars. 

\subsection{Testing Scenarios}
\subsubsection{Artificial Data}
We design six scenarios as summarized in Table \ref{tab:scenario} to test our algorithms. Each set of scenarios contains 40 problem instances.
\begin{itemize}[leftmargin=*]
	\item Grid world: we design a $20\times20$ grid world. Each grid is a $1 km\times 1km$ square area. The grid world are defined into three types: intense, peripheral and empty, with the percentage of each grid type equals 5\%,15\%,80\% respectively. As shown in Table \ref{tab:scenario}, the type of each grid is fixed initially in all instances except for the \emph{Random Grid}, which has the grid type randomly generated based on the probability.
	\item Time horizon: time horizon equals 480 minutes and is equally partitioned into eight periods as shown in Table \ref{tab:ArrivalRate}.
	\item Couriers: Three sets of courier number are conducted in our experiments to evaluate the scalability of \emph{MARL} model. Each courier has a moving speed of 0.5 km/minute. All couriers departure from the central gird (10,10).
	\item Pickup request: according to Table \ref{tab:ArrivalRate}, we generate pickup requests for each grid and each time period in a Poisson process and set the current time as $\underline{w_i}$ for request $r_i$. The location of each request is uniformly generated within the corresponding grid.

	\item Degree of dynamism (DOD): to evaluate the impact of information known in advance to different algorithms, we adjust the number of pickup request known at time 0. For example, if the value of DOD equals 90\% as shown in Table \ref{tab:scenario}, we randomly choose $10\%$  of the total generated requests and set their $t_i$ to time 0. 
\end{itemize}

\begin{table}[]
\scriptsize
	\caption{Arrival Rate Matrix}
	\label{tab:ArrivalRate}
	\begin{tabular}{@{}ccccccccc@{}}
		\toprule
		Grid Type      & \multicolumn{8}{c}{Customer Arrival Rate}                     \\ \midrule
		intense    & 0.05 & 0.00 & 0.00 & 0.10 & 0.04 & 0.00 & 0.00 & 0.05 \\
		peripheral & 0.01 & 0.06 & 0.01 & 0.01 & 0.01 & 0.06 & 0.05 & 0.01 \\ \bottomrule
	\end{tabular}
\end{table}

\begin{table}
	\renewcommand\arraystretch{1.3}
	\tiny
	\newcommand{\tabincell}[2]{\begin{tabular}{@{}#1@{}}#2\end{tabular}}
	\caption{Scenario settings}
	\label{tab:scenario}
	\begin{tabular}{c|c|c|c|c|c}
		\hline
		\multirow{2}{*}{\tabincell{c}{ Scenario\\ Type}} &
		\multirow{2}{*}{\tabincell{c}{ Time \\ Window}} & 
		\multirow{2}{*}{\tabincell{c}{Degree of \\ Dynamism } }  & 
		\multirow{2}{*}{\tabincell{c}{Zone \\ Distribution }}  & 
		\multirow{2}{*}{\tabincell{c}{Courier \\ Number \\}} & 
		\multirow{2}{*}{\tabincell{c}{Customer \\ Number \\}} \\   & & & & &  \\  \hline
		{\emph{Base World}} & 60 min & 90\% &  fixed & 10 & 1000\\ 
		{\emph{Median World}} & 60 min & 90\% & fixed  & 30 & 3000 \\ 
		{\emph{Large World}} & 60 min & 90\% & fixed  & 100 & 15000 \\ 
		{\emph{Small TW}} & 20 min & 90\% & fixed  & 10 & 1000 \\ 
		{\emph{Low Dynamism}} & 60 min & 50\% & fixed  & 10 & 1000 \\ 
		{\emph{Random Grid}} & 60 min & 90\% & random  & 10 & 1000 \\\hline
	\end{tabular}
\end{table}


\subsubsection{Real-world Data}
We use the real-world data provided by the courier pickup services of Alibaba in the city of Hangzhou. We select an area of 20km $\times$ 20km and divide the area into 20${\times}$20 grids. Different to the artificial data, we adopt real distances between grids that are provided by GaoDe map api\footnote{https://lbs.amap.com/}. We choose 60 continuous days of data and use the first 30 days' data as training set and the second half as the test set for later experiments.

\section{Experiment Design} 
\subsection{Compared Algorithms}
The \emph{MARL} model is compared to the following dispatching algorithms:
\begin{itemize}[leftmargin=*]
\item \emph{Random}: This method does not require any information about the current state. It randomly selects a neighborhood grid and a service period from the action space $\mathcal{A}$.

\item \emph{GHAV}(Greedy to the highest absolute value): The highest absolute value dispatcher selects a grid with a service period that currently has the highest value per unit time from the action space $\mathcal{A}$.

\item \emph{GHEP}(Greedy to the highest expected profit): The highest expected profit dispatcher selects a grid and a service period that has the highest expected profit related to the requested courier. We define the expected profit of each candidate action by considering the traveling time to the target grid and pre-calling the route planning solver to identify possible gains after arriving the target grid.

\item \emph{MBM}(Maximum bipartite matching): Maximum bipartite matching dispatcher tries to maximize the total score for all couriers that will be available in 20 minutes. More particularly, in the bipartite graph, there are two disjoint subsets. One sub-graph includes available couriers and the other includes the cartesian product of candidate grids and service periods. We calculate the expected profit of each candidate courier to each grid with each service period. Then a maximum optimal matching problem is solved and the current courier is dispatched. 
\end{itemize}

\subsection{Experimental Settings}
In our experiment, training step lasts for 10000 episodes, and the performance on both the train set and test set are reported. The similar network structure is applied for both value function and policy network, which is parameterized by one fully-connected hidden layer of 200 units with ReLU. After each episode, the sample of transition tuples is written into the memory buffer, which has the size of 20000. In learning step, iteration number $N_{1}$ and $N_{2}$ is set to 10, with the batch size of 1024. Adam optimizer is applied, with the learning of $5\times10^{-4}$.  In considering the collaborative revenue of the whole fleet, performance with the reward reshaping weight $\alpha$ of 0.1, 0.5 and 0.9 has been compared, and $\alpha$ value of 0.5 obtains the highest score with a slight advantage of 0.16\% and 0.67\% against other two settings, and therefore $\alpha$ value of 0.5 is applied in the following discussion. In the PPO setting, clipping value $\epsilon$ is set to 0.2. 

\subsection{Performance Comparison}
As shown in Table \ref{tab:results1}, the performance of each dispatching strategy is presented. The score is calculated by the percentage of achieved price over the total price of all orders. For the proposed \emph{MARL}, we feed different information to the presentation of states. At basic information level, the dispatch system only obtains the current information of customers and couriers, which is denoted as \emph{MARL-B} in Table \ref{tab:results1}. Furthermore, we add an additional expected profit channel that is the same information used in \emph{GHEP} and \emph{MBM}, which is denoted as \emph{MARL-EP} 



On both the base-world and real-world dataset, \emph{Random} is the worst strategy, which could only get about 23\% score of all orders appeared in the whole day. By comparing the performance of \emph{GHAV} and \emph{GHEP}, the expected value function does help couriers to identify the relative best target grid more accurately. This pattern is also shown in the results of \emph{MARL-B} and \emph{MARL-EP}. \emph{MBM} has a significant increase in score compared to \emph{GHAV} and \emph{GHEP}. This is attributed to the consideration of global resource and demand allocation when making a single dispatching decision. \emph{MARL-EP} gives the best achievement in score in all scenarios. These results provide good evidence that \emph{MARL} has the learning ability of sophisticated decision process. 

 

\begin{table}
	\renewcommand\arraystretch{1.3}
	\scriptsize
	\newcommand{\tabincell}[2]{\begin{tabular}{@{}#1@{}}#2\end{tabular}}
	\caption{Comparison of performance}
	\label{tab:results1} 
\begin{tabular}{c|c|c|c|c}
	\hline
	\multirow{2}{*}{Strategy} & \multicolumn{2}{c|}{Base World}     & \multicolumn{2}{c}{Real World}     \\ \cline{2-5} 
	& train set        & test set         & train set        & test set         \\ \hline
	\textit{Random}           & 23.35\%          & 22.99\%          & 24.56\%          & 27.78\%          \\ \hline
	\textit{GHAV}             & 74.17\%          & 74.07\%          & 71.24\%          & 74.86\%          \\ \hline
	\textit{GHEP}             & 75.12\%          & 75.51\%          & 73.47\%          & 75.30\%          \\ \hline
	\textit{MBM}              & 81.13\%          & 80.77\%          & 76.46\%          & 75.93\%          \\ \hline
	\textit{MARL-B}             & 82.80\%          & 83.00\%          & 77.86\%          & 77.96\%          \\ \hline
	\textit{MARL-EP}          & \textbf{\footnotesize 84.27\%} & \textbf{\footnotesize 84.45\%} & \textbf{\footnotesize 79.16\%} & \textbf{\footnotesize 79.21\%} \\ \hline
\end{tabular}
\end{table}

\subsection{Agent Scale}
In discussion of applicability of the \emph{MARL} in urban scale level, \emph{Median World} with 30 couriers, 3000 customers and \emph{Large World} with 100 couriers, 15000 customers are studied, which is significantly larger than the problem instances studied in the MDP modelling group (e.g. \cite{Novoa2009,Goodson2013,Ulmer2016,Ulmer2017,Ulmer2018}). The detailed performance on all scenarios is listed in Table \ref{tab:results_scale}. The \emph{Median World} has relatively sufficient number of couriers. \emph{MARL-EP} presents about 1\% improvement against the \emph{MBM}. 

In the \emph{Large World},  we firstly observe the better performance of \emph{GHAV} compared to \emph{GHEP}. This is due to the mutual influence between a large number of agents leads to a rapid change of grid values, thus the expected value function fails to give a correct evaluation of the grid value. Nevertheless, \emph{MARL-EP} still gets the best performance, having at least 3\% increase against all human-designed methods.


\begin{table}[H]
	\renewcommand\arraystretch{1.3}
	\scriptsize
	\newcommand{\tabincell}[2]{\begin{tabular}{@{}#1@{}}#2\end{tabular}}
	\caption{Performance on extended agent scale}
	\label{tab:results_scale} 
	\begin{tabular}{c|c|c|c|c}
		\hline
		\multirow{2}{*}{Strategy} & \multicolumn{2}{c|}{Median World}   & \multicolumn{2}{c}{Large World}    \\ \cline{2-5} 
		& train set        & test set         & train set        & test set         \\ \hline
		\textit{Random}           & 35.75\%          & 34.77\%          & 27.57\%          & 27.94\%          \\ \hline
		\textit{GHAV}             & 82.00\%          & 81.46\%          & 65.75\%          & 64.87\%          \\ \hline
		\textit{GHEP}             & 84.09\%          & 84.27\%          & 64.44\%          & 64.22\%          \\ \hline
		\textit{MBM}              & 91.37\%          & 91.45\%          & 74.71\%          & 74.62\%          \\ \hline
		\textit{MARL-B}             & 91.87\%          & 91.99\%          & 77.77\%          & 77.85\%          \\ \hline
		\textit{MARL-EP}          & \textbf{\footnotesize 92.45\%} & \textbf{\footnotesize 92.58\%} & \textbf{\footnotesize 78.47\%} & \textbf{\footnotesize 78.51\%} \\ \hline
	\end{tabular}
\end{table}

\subsection{Model Generality}
Typically, reinforcement learning is trained on a specific environment setting, the generalization capability of RL model is beyond the scope in many problems. However, due to the uncertainty and unpredictability of online situation, it is worth discussing if there exists certain transferability of the dispatching strategy acquired. On this topic, the generality and extensibility of our model is discussed on types of 3 datasets derived from the base scenario, mainly varying on the aspects of customer service time window, the degree of dynamic orders, and customer zone distribution. In scenario \emph{Small TW}, the time window of available service time is reduced to 20 minutes, which require a more real-time response to new appear customers. In scenario \emph{Low Dynamism}, the ratio of dynamic orders reduced to 50\%, and in scenario \emph{Random Grid}, the zone distributions are randomly shuffled. 

\begin{table}[]
	\renewcommand\arraystretch{1.3}
	\scriptsize
	\newcommand{\tabincell}[2]{\begin{tabular}{@{}#1@{}}#2\end{tabular}}
	\caption{Performance on different scenario}
	\label{tab:results3} 
	\begin{tabular}{c|c|c|c}
		\hline
		\multirow{2}{*}{Strategy} & \multicolumn{3}{c}{Dataset}                           \\ \cline{2-4} 
		& Small TW         & Low Dynamism     & Random Grid      \\ \hline
		\textit{Random}           & 25.21\%          & 31.01\%          & 23.47\%          \\ \hline
		\textit{GHAV}             & 64.98\%          & 73.49\%          & 72.85\%          \\ \hline
		\textit{GHEP}             & 68.15\%          & 77.58\%          & 76.48\%          \\ \hline
		\textit{MBM}              & 68.76\%          & 79.00\%          & 80.68\%          \\ \hline
		\textit{MARL-B}             & 71.62\%          & 81.60\%          & 81.34\%          \\ \hline
		\textit{MARL-EP}          & \textbf{\footnotesize 72.50\%} & \textbf{\footnotesize 81.51\%} & \textbf{\footnotesize 82.71\%} \\ \hline
	\end{tabular}
\end{table}	

The detailed performance on all scenarios is listed in Table \ref{tab:results3}. \emph{MARL-EP} still presents a significant advantage over \emph{GHAV}, \emph{GHEP} and \emph{MBM} on all the scenarios, though the gap is less obvious than on the trained scenario. The resultes proves that \emph{MARL} is capable of handling unseen dynamics in real scenarios.

\subsection{Imitation Learning}
\begin{figure}[ht]
	\center
	\includegraphics[scale = 0.42]{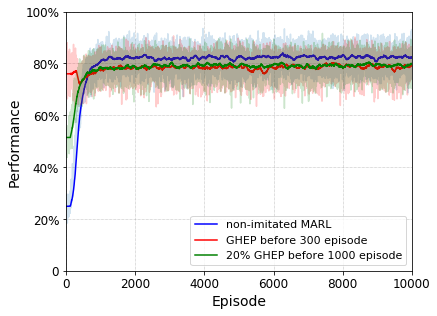} 
	\caption{Performance with imitated strategy from GHEP dispatcher}
	\label{fig:imitation}
\end{figure}

The major shortcoming of reinforcement learning lies in its low sampling efficiency, leading to long warm-up phase during the learning. In many reinforcement learning tasks, if high-quality samples can be generated through historical data or strategies, then the policy might be trained quickly through imitation learning, by using the privilege of past experience and expert strategy. 
In our experiment, \emph{GHEP} is selected as the imitated objective for \emph{MARL-EP}. Two attempts have been conducted, specifically all \emph{GHEP} strategy before the $300^{th}$ episode, and 20\% probability of \emph{GHEP} strategy before $1000^{th}$ episode. The performance is listed in figure \ref{fig:imitation}. It is obvious that imitation learning only accelerate the initial process of the learning curve, while the long-term performance is less favorable than the none-imitated model. The possible explanation is that, reinforcement learning method in this dispatching problem can be easily guided into a local optimum by the imitated strategy. Exploration by self-discovery is quite important in the RL framework, excessive exploitation of expert experience might block the opportunity to learn the best strategy from a long-term perspective. 

\subsection{Cooperation Discussion}
In a multi-agent system, it is likely that not all agents are identically under control, especially in the real business scenario. In the courier dispatch system, couriers might be managed and controlled by different dispatch platforms, and it is worth discussing if \emph{MARL} is capable of learning dispatching strategy in a cooperative environment. In the following discussion, the experiment of four fleets is discussed, each with the courier number of 10. In fleet 1, all 10 couriers share the same policy. In the other three fleets, two sets of dispatching policies are adopted, with each policy controls the implementation of five couriers. In fleet 2, the first 5 couriers use \emph{MARL-EP} policy, and the other 5 couriers use \emph{GHEP} policy. In fleet 3, the first 5 couriers use \emph{MARL-EP} policy, and the other 5 couriers use \emph{Random} policy. In fleet 4, each group of five couriers uses an independent learning environment, in which samples are independently collected and model is separately trained. Due to the homogeneous nature of all couriers, the performance obtained by fleet 4 is approximately equal to fleet 1. However, the learning progress is a bit slower in fleet 4 due to the smaller sample size of each model. In fleet 2, the total performance is a bit lower than the pure \emph{MARL-EP} strategy. At the beginning stage of the learning phase, \emph{GHEP} gets a quite high score, while \emph{MARL-EP} is roughly random. However, as the training process proceeds, the performance of \emph{MARL-EP} gradually grows and surpasses the \emph{GHEP} group at the $1000^{th}$ episode, and the final performance of \emph{MARL-EP} group is 13\% higher than the 
\emph{GHEP} group. In fleet 3, the five couriers in \emph{MARL-EP} group obtains the highest group revenue by making up the loss caused by the inefficient random strategy.

\begin{figure}[ht]
	\center
	\includegraphics[scale = 0.45]{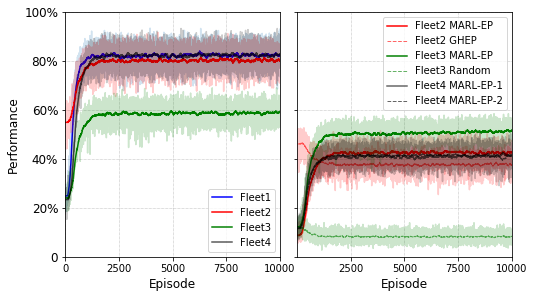}
	\caption{Performance with two set of strategy}
	\label{fig:collaborative}
\end{figure}

\subsection{Strategy Analysis}
\begin{figure}[ht]
	\center
	\includegraphics[width=\linewidth]{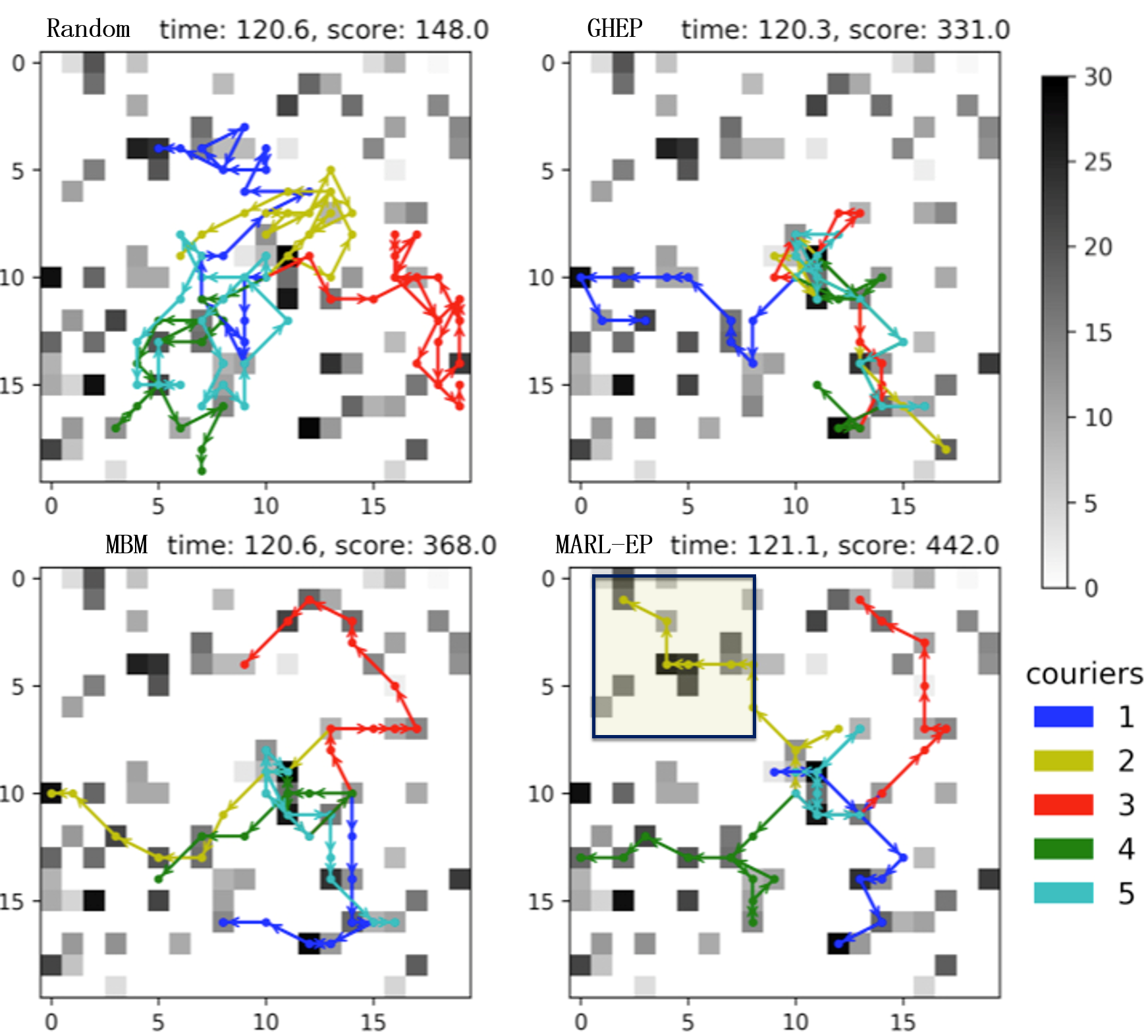}
	\caption{Trajectory of 5 couriers in the first two hours of Scenario 1}
	\label{fig:analysis_toy}
\end{figure}

In this section, we aim to gain an insight into the derived dispatching policy by \emph{MARL-EP} due to its outstanding performance. We analyze the trajectory of each courier as shown in Figure \ref{fig:analysis_toy} from dataset scenario 1. In each subgraph, the color shown in the grid represents the cumulative price of appeared pickup requests in the last two hours, where the light color represents low price and dark color represents high price. Further, we set up the same initial state to all tested scenarios. Note that the pickup requests in our problem settings have hard constraints of service time window and disappeared time, which leads to a high requirement of precision spatial and temporal matching. Interesting behavior can be seen from figure \ref{fig:analysis_toy}. a) The random dispatcher generates the longest walk trace and shortest pick-time that actually gain value. b) The \emph{GHEP} shows no cooperation between couriers. This strategy leads to agents' competition in local areas, which results in low overall revenue. c) The \emph{MBM} considers cooperation when there are enough pickup requests showing up. However, it is difficult to make an advisable decision during the period when there are not enough requests. d) In comparison, the \emph{MARL-EP} dispatcher is able to capture the changes of request distribution over time. Take the area highlighted in \emph{MARL-EP} of figure \ref{fig:analysis_toy} as an evidence, the myopic strategies fail to explore this area due to its isolated and remote location. Only pre-dispatching at the early time could capture the orders of this area. 

\begin{figure}[ht]
\center
\includegraphics[width=\linewidth]{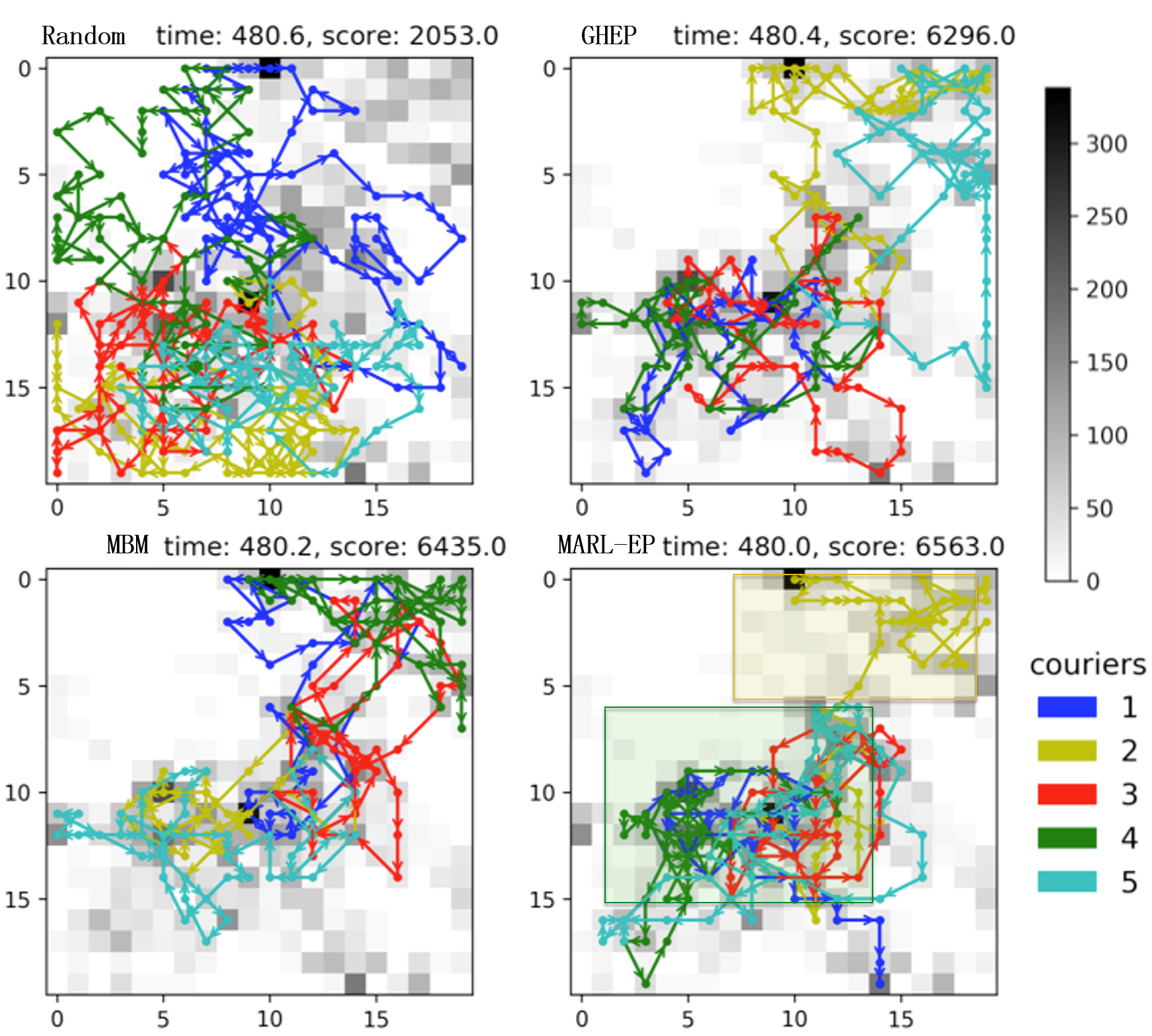}
\caption{Trajectory of 5 couriers in the whole day on $Aug.5^{th}  \ 2018$ }
\label{fig:analysis_real}
\end{figure}

Figure \ref{fig:analysis_real} presents the dispatching paths of different methods given real-world data of Hangzhou city over 8 hours. Similar analysis and conclusion can be drawn. In addition, better supply and demand matching between north and center area over spatial and temporal space is observed in \emph{MARL-EP}.

\section{Conclusion}
In this paper, we formulate the courier dispatching problem as a Markov decision process and use a multi-agent reinforcement learning method to solve this problem. We propose an effective dispatching algorithm with decentralized control and reward shaping to encourage cooperation among agents. The results from experiments on both the artificial dataset and the real-world dataset show that the MARL achieves significant improvement over the human-designed dispatching policies. Our method is capable of capturing intrinsic patterns among data and make reasonable decisions from a long-term perspective. Model generality is confirmed on the unseen scenarios beyond the training dataset, mainly on the scope of the customer time window, dynamic ratio, and zone distribution variation. Moreover, we show the scalability of the proposed method on significantly larger data sets compared to existing literature.

In future research, we will focus on investigation of more effective network architecture and learning algorithms. Meanwhile, it is beneficial to apply our proposed method to more interesting combinatorial optimization problems in the domain of logistics and online scheduling systems.

\begin{acks}
The authors would like to thank the anonymous referees for their valuable comments and helpful suggestions. The authors would also like to show our gratitude to Prof. Yang Yu from School of Artificial Intelligence of Nanjing University for his valuable advices. 
\end{acks}


\bibliographystyle{ACM-Reference-Format}  
\balance  
\bibliography{aamas19_539}  

\end{document}